\documentclass[conference]{IEEEtran}
\IEEEoverridecommandlockouts
\usepackage{cite}
\usepackage{amsmath,amssymb,amsfonts}
\usepackage{physics}
\usepackage{algorithmic}
\usepackage{algorithm2e}
\usepackage{graphicx}
\usepackage{siunitx,booktabs}
\usepackage{textcomp}
\usepackage{xcolor}
\def\BibTeX{{\rm B\kern-.05em{\sc i\kern-.025em b}\kern-.08em
    T\kern-.1667em\lower.7ex\hbox{E}\kern-.125emX}}
    
\usepackage[colorlinks=true,citecolor=blue,linkcolor=blue]{hyperref}

\usepackage[scale=2]{ccicons}

\usepackage{mathtools}
\usepackage{siunitx}
\usepackage[utf8]{inputenc}

\usepackage{listings}
\usepackage{xcolor}

\definecolor{codegreen}{rgb}{0,0.6,0}
\definecolor{codegray}{rgb}{0.5,0.5,0.5}
\definecolor{codepurple}{rgb}{0.58,0,0.82}
\definecolor{backcolour}{rgb}{0.95,0.95,0.92}

\lstdefinestyle{mystyle}{
    backgroundcolor=\color{backcolour},   
    commentstyle=\color{codegreen},
    keywordstyle=\color{magenta},
    numberstyle=\tiny\color{codegray},
    stringstyle=\color{codepurple},
    basicstyle=\ttfamily\footnotesize,
    breakatwhitespace=false,         
    breaklines=true,                 
    captionpos=b,                    
    keepspaces=true,                 
    numbers=left,                    
    numbersep=5pt,                  
    showspaces=false,                
    showstringspaces=false,
    showtabs=false,                  
    tabsize=2
}
\title{Deep Neural Network-Based Sign Language Recognition: A Comprehensive Approach Using Transfer Learning with Explainability}
\makeatletter
\newcommand{\linebreakand}{%
  \end{@IEEEauthorhalign}
  \hfill\mbox{}\par
  \mbox{}\hfill\begin{@IEEEauthorhalign}
}
\makeatother

\author{
  \IEEEauthorblockN{\textsuperscript{} A. E. M Ridwan\IEEEauthorrefmark{1}}
  \IEEEauthorblockA{\textit{Computer Science and Engineering} \\
    \textit{BRAC University}\\
    Dhaka, Bangladesh \\
    a.e.m.ridwan@g.bracu.ac.bd}
  \and
  \IEEEauthorblockN{\textsuperscript{}Mushfiqul Islam Chowdhury\IEEEauthorrefmark{1}}
  \IEEEauthorblockA{\textit{Computer Science and Engineering} \\
    \textit{BRAC University}\\
    Dhaka, Bangladesh \\
    mushfiqul.islam.chowdhury@g.bracu.ac.bd}

  \linebreakand
  \IEEEauthorblockN{\textsuperscript{} Mekhala Mariam Mary}
  \IEEEauthorblockA{\textit{Computer Science and Engineering} \\
    \textit{BRAC University}\\
    Dhaka, Bangladesh \\
    mekhalamariam@gmail.com}
  \and
  \IEEEauthorblockN{\textsuperscript{} Md Tahmid Chowdhury Abir}
  \IEEEauthorblockA{\textit{Computer Science and Engineering} \\
    \textit{BRAC University}\\
    Dhaka, Bangladesh \\
    md.tahmid.chowdhury.abir@g.bracu.ac.bd}
    \thanks{\IEEEauthorrefmark{1}These authors contributed equally to this work.}
}

\usepackage[pscoord]{eso-pic}
\newcommand{\placetextbox}[3]{
 \setbox0=\hbox{#3}
 \AddToShipoutPictureFG*{ \put(\LenToUnit{#1\paperwidth},\LenToUnit{#2\paperheight}){\vtop{{\null}\makebox[0pt][c]{#3}}}
 }
 }
\placetextbox{0.5}{0.055}{\small{979-8-3503-7550-3/24/\$31.00 ©2024 IEEE}}

\begin{document}

\maketitle

\begin{abstract}
 To promote inclusion and ensuring effective communication for those who rely on sign language as their main form of communication, sign language recognition (SLR) is crucial. Sign language recognition (SLR) seamlessly incorporates with diverse technology, enhancing accessibility for the deaf community by facilitating their use of digital platforms, video calls, and communication devices. To effectively solve this problem, we suggest a novel solution that uses a deep neural network to fully automate sign language recognition. This methodology integrates sophisticated preprocessing methodologies to optimise the overall performance. The architectures resnet, inception, xception, and vgg are utilised to selectively categorise images of sign language. We prepared a DNN architecture and merged it with the pre-processing architectures.  In the post-processing phase, we utilised the SHAP deep explainer, which is based on cooperative game theory, to quantify the influence of specific features on the output of a machine learning model. Bhutanese-Sign-Language (BSL) dataset was used for training and testing the suggested technique. 
 While training on Bhutanese-Sign-Language (BSL) dataset, overall ResNet50 with the DNN model performed better accuracy which is 98.90\%. Our model's ability to provide informational clarity was assessed using the SHAP (SHapley Additive exPlanations) method. In part to its considerable robustness and reliability, the proposed methodological approach can be used to develop a fully automated system for sign language recognition.
 
 %Our modified Vgg16   obtains trianing accuracy of 98.53\% , test accuracy of 92.04\%, precision of 92.33\%, F1 score of 92.04\%, and recall of 92.04\%. Our proposed ResNet50's training accuracy, test accuracy, precision, F1 score, and recall are 98.90\%, 90.24\%, 90.04\%, 90.24\%, and 90.24\% respectively. Our modified Inceptionv3 model provides 95.47\% training accuracy, 84.98\% test accuracy, 84.85\% precision, 84.98\% F1 score, and 84.98\% recall. Separate training on our modified Xception's training accuracy rates of 98.80\%, test accuracy rates of 87.75\%, precision rates of 87.59\%, F1 score of 87.75\%, and recall rates of 87.75\%.

\end{abstract}


\section{Introduction}
Recognizing and interpreting hand movements, such as sign language, has become a crucial technology in the field of human-computer interaction with a wide range of applications. Hand gesture recognition (HGR) encompasses machine vision, signal analysis, pattern identification, and human-computer interaction \cite{intro1}. It's sub-branch Sign language is a sophisticated and nuanced mode of communication that surpasses the limitations of verbal language, especially for those who are hard of hearing or deaf. The transmission of meaning is dependent on the use of visual-manual gestures, facial emotions, and body movements. 

Complex forms of verbal communication, such as American Sign Language (ASL) and British Sign Language (BSL), play a vital role in facilitating communication for those who are deaf or have hearing impairments. Sign language caters to the unique linguistic needs of various communities and enables complex and culturally diverse communication by employing its inherent syntax and vocabulary, going beyond mere manual motions. This introduction recognises the legitimacy of sign language and its essential use in educational, public, and legal contexts. Skilled interpreters are responsible for upholding this practice, as they are dedicated to facilitating communication and promoting inclusively for those with diverse language requirements.

This study focuses on the specific topic of static sign language recognition (SLR) as its major subject since the deaf community frequently employs the capacity to read hand movements that match to letters and numbers as their primary form of communication. Recently, deep learning has become increasingly popular as a technique for use in the field of hand gesture recognition (HGR). 
%This is accomplished by utilizing tools like stacking denoising autoencoders (SDAEs) \cite{intro2}, convolutional neural networks (CNNs) \cite{intro2} -\cite{intro5}, and most notably recurrent neural networks (RNNs) \cite{intro6}.

This paper made the following contribution, which are summarized below:
\begin{itemize}
    \item Architecting a Deep Neural Network (DNN) for Classification on Bhutanese-Sign-Language (BSL)
    
    \item Evaluating the Efficacy of Various Transfer Learning Models as Feature Extractors for our Architectured Model with accuracy of 98.90\% for  ResNet50,  95.47\%  for Inceptionv3, and 98.80\% for Xception.
    
    \item Elucidating Model Outputs by employing Xplanable AI to draw Insights from Training Procedures and Data Characteristics.
\end{itemize}
  
In Section \ref{related work}, we briefly examine the relevant literature. The experimental materials and procedures are outlined in Section \ref{methodology1}. The accessible datasets are also described in section \ref{dataset}. Pre-processing the data is shown in Section \ref{preprocessing}. Section \ref{our model} includes a concise summary of the proposed architecture, while Section \ref{eva matrix} details the evaluation matrix. Experimental results are presented in Section \ref{result} and \ref{Comparative Analysis} where the existing study on different datasets are compared with the efficiency of our built network. Section \ref{conclude} provides conclusion.

\section{Literature Review}\label{related work}

This section will provide a concise overview of the research work pertaining to our study.\newline
Ozcan et al. \cite{review1} suggested using a transfer learning-based Convolutional Neural Network (CNN) to identify human motions. During their research, they achieved 98.40\% success rate for ABC tuned-CNN and 98.09\% success rate for Thomas Moeslund’s gesture recognition dataset. Abeer et al. \cite{review2} converted grayscale images to three-channel images for hand gesture recognition. They implemented ResNet50 \cite{review9} and MobileNetV2 \cite{review11} architectures to obtain 97\% accuracy. Another study by Karsh et al. \cite{review3} introduced a two-phase deep learning-based technique dubbed modified inception v3 (mIV3Net) to overcome the problems of complicated backdrops and inter-class similarities in current hand gesture recognition (HGR) systems. Improved gesture recognition across five different datasets is shown using the proposed method, which involves the exploration of different transfer learning techniques and the optimisation of model performance through hyper-parameter tuning. In a similar study, Kothadiya et al. \cite{review4} applied Attention-based ensemble learning to improve sign language recognition and introduced an architecture featuring eXplainable Artificial Intelligence (XAI) for sign language recognition. They also presented actual evidence showcasing the interpretability and decision-oriented aspect of the suggested methodology utilizing Explainable AI \cite{review10} techniques. Beser et al. \cite{review5} utilized a technique for recognising sign language based on capsule networks (CapsNet) \cite{review6}. Changes were made to CapsNet's batch size, convolution layer, filter size, and augmentation of data. In their experimental experiments, they found an average success rate of 94.2\% using the sign language digits dataset. Das et al. \cite{review7} investigates the effectiveness of sign language recognition with a particular focus on ASL (American Sign Language). Using retraining and testing on an Inception v3-based convolutional neural network \cite{review8}, the research achieves a validation accuracy of greater than 90\%. Wu et al. \cite{review12}  addressed the underutilization of the Xception model \cite{review13} in scene image classification and proposes an Xception-based transfer learning strategy, demonstrating its superior performance in comparison to Inception-V3. Improved model performance and practicality are the results of Xception-based transfer learning, which the study proves to be effective in scene picture categorization and which also stands out for its generalizability, robustness, and capacity to avoid overfitting. Feng et al. \cite{review14} proposed the critical nature of gesture recognition technology in human-computer interaction (HCI) is significantly hampered by the fact that it is highly susceptible to illumination conditions. By extracting HOG features from gestures and classifying them with Support Vector Machines (SVMs), the paper addresses the issue and achieves high recognition rates even in varying illumination conditions. Islam et al. \cite{review15} investigated the role that hand gestures have in human-computer interaction and put forth a Convolutional Neural Network (CNN)-based static hand gesture identification technique. The model obtains a greater accuracy of 97.12\% with data augmentation approaches than it does without augmentation, proving the efficacy of CNNs in identifying intricate and non-linear correlations among images. 

Using a scalable strategy that integrates preprocessing, model training, and postprocessing, we provide a deep neural network-based approach to fully automated sign language recognition (SLR). We fed the dataset into the transfer learning (TL) model without its top layer to extract features, with all layers purposely frozen to maintain pre-trained weights. Later, for adaptation, a custom classification layer was introduced, reshaping the picture data using an input tensor. We employed dense layers with dropout for regularisation. ResNet, Inception V3, Xception, and VGG architectures and categorical cross-entropy loss, Adam optimizer, and evaluation metrics were used to enhance efficiency of our method which is confirmed by thorough assessments on Bhutanese-Sign-Language (BSL) datasets. At the end of the thorough investigation was carried out to evaluate the model's ability to clarify information using the SHAP technique. 

\section{Proposed Methodology}\label{methodology1}
\subsection{Workflow Diagram}

The study's overarching goal is to demonstrate that deep learning techniques, and more especially transfer learning as a feature extraction method, can successfully detect sign language. 
Subsequently, an examination was conducted to assess the elucidation capabilities of the model utilising the SHAP (SHapley Additive exPlanations) technique.

\begin{figure}[t]
\centerline{\includegraphics[width=3.5in, height=6in]{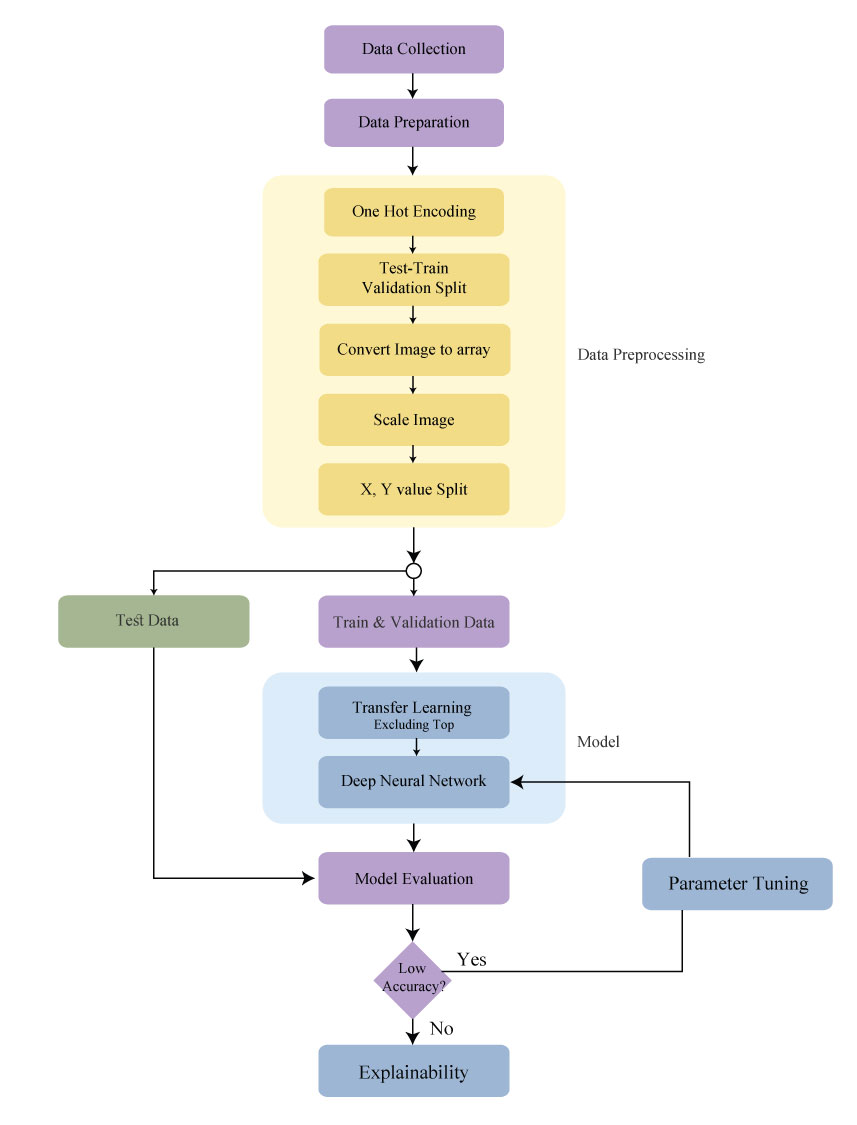}}

\caption{Work Flow Diagram}
\label{Work Flow Diagram}
\end{figure}

\subsection{Methodology}\label{}
% Initially, following the preparation of the dataset, we proceeded to do feature extraction utilizing four distinct techniques, namely vgg16, inceptionv3, resnet16, and exception. Subsequently, a classification task was undertaken employing a modified variant of Deep Neural Network (DNN) in order to achieve optimal accuracy. Subsequently, the SHAP methodology was employed to obtain the explainability of the model.

This section presents a detailed breakdown of the various components that make up the overall procedure of the study. The Bhutanese-Sign-Language (BSL) datasets were employed in the suggested frameworks, namely vgg16 \cite{vgg}, inceptionv3 \cite{ review9}, resnet16 \cite{review8}, and exeption \cite{review13}. We inputted the dataset into the transfer learning (TL) model, excluding its top layer, in order to extract features. Figure \ref{Work Flow Diagram} illustrates that we intentionally kept all layers frozen to preserve the pre-trained weights. Subsequently, a specialised classification layer was implemented to modify the visual data by utilising an input tensor. We utilised evaluation measures to optimise the efficiency of our approach.

\section{Dataset Description}\label{dataset}

This is Bhutanese-Sign-Language (BSL) \cite{dataset} of 10 digits. Multiple actors' performances were captured on video before frames were extracted. 20,000 images with varying resolutions (2000 for each class) made up the BSL digits dataset. 

\begin{figure*}[t]
\centerline{\includegraphics[width=6.08in, height=3.74in]{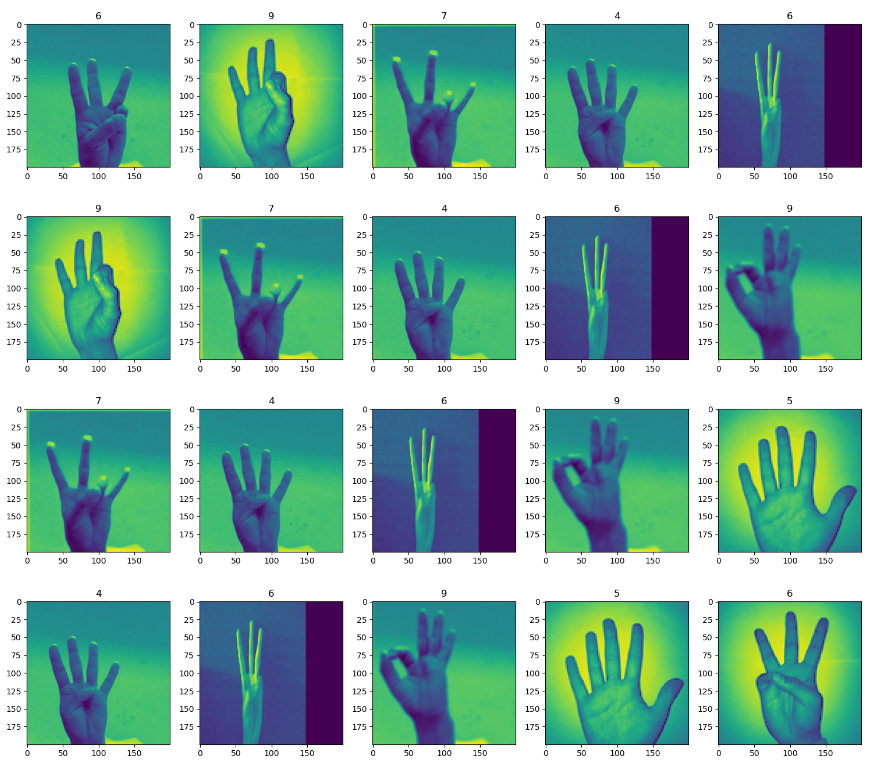}}

\caption{Representation of Bhutanese Sign Language Dataset}
\label{Representation of Bhutanese Sign Language Dataset}
\end{figure*}

\section{Data Preparation and Preprocessing}\label{preprocessing}

The initial image dimensions were 200x200 pixels, which is presented in Figure \ref{Representation of Bhutanese Sign Language Dataset}. After a comprehensive examination of the images, it was determined that they should be resized to 75x75 pixels. This modification was implemented to maintain uniformity in input image dimensions throughout the dataset, thereby preserving all relevant data.

Our dataset consists of ten distinct output classes, each of which was encoded using one-hot encoding.  One typical approach, called label encoding, uses numerical values to represent categorical labels, but it can be biased because it suggests a hierarchy where none exists. To solve this problem, we can utilise one-hot encoding to give numerical values to categorical variables without assuming any sorting. We utilised one hot encoding to classify the data in our dataset. For instance, we need to figure out if the picture of the number 3 represents a 0 or a 1, and similarly for the image of the number 4.

Due to the resource-intensive nature of image datasets, including training, testing, and explanation stages, a strategy was implemented to control memory consumption. Initially, the dataset was divided into distinct sets for training, testing, and validation, followed by data processing in accordance with subsequent procedural requirements.

During the training phase, the train and validation sets were transformed into arrays and then normalized to improve information extraction. These normalized arrays were used to train models. For evaluating and explaining the results of our models, the same sequence of steps was maintained.

\section{Proposed Achitecture}\label{our model}
Starting with the transfer learning (TL), we passed the dataset to the TL model without the top layer for feature extraction. We intentionally freeze all layers to prevent the pre-trained weights from being updated during training, we added the custom classification layer later.  We  defined the input tensor shape as (75, 75, 3) to reshape our image data.
\begin{figure*}[t]
\centerline{\includegraphics[width=7.33in, height=2in]{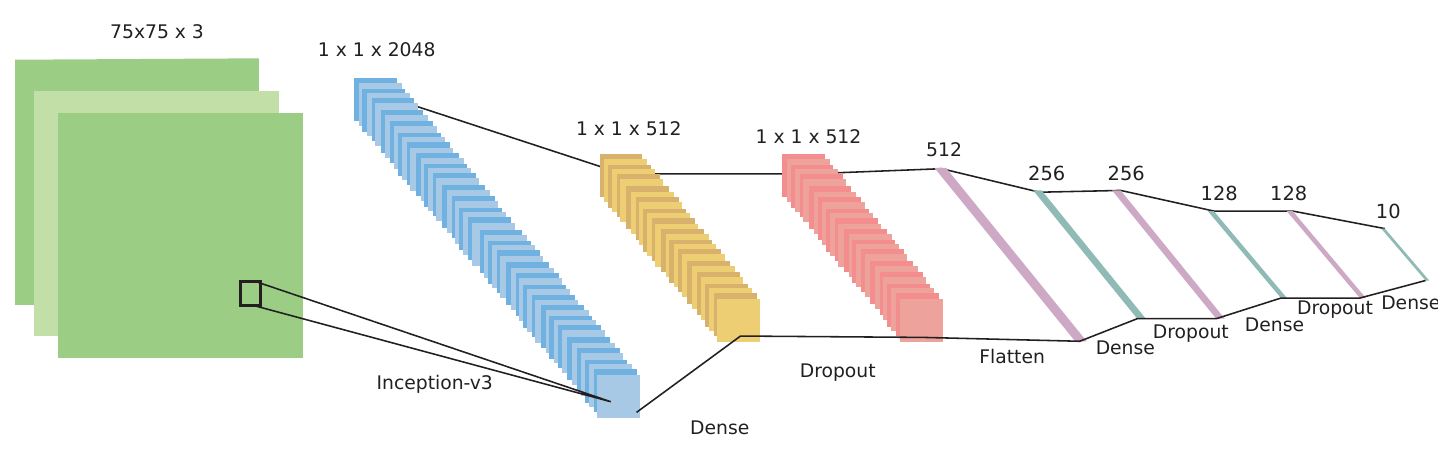}}

\caption{DNN-inception v3}
\label{DNN-inception}
\end{figure*}

\begin{figure*}[t]
\centerline{\includegraphics[width=7.33in, height=2in]{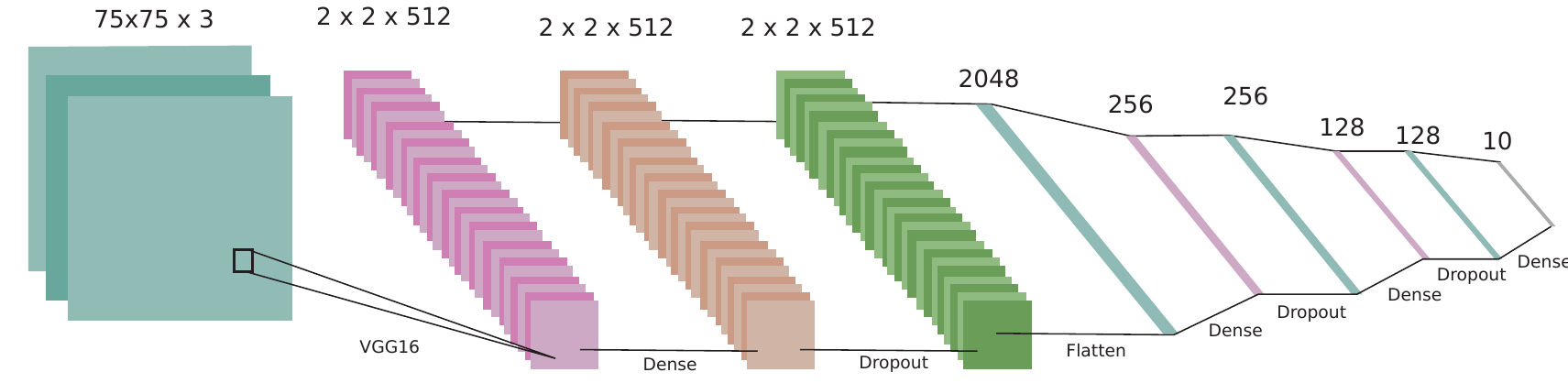}}

\caption{DNN-VGG16}
\label{DNN-VGG}
\end{figure*}

\begin{table*}[t]
\centering
\begin{tabular}{c  c}
\hline
Transfer learning & Pre-trained on Bhutanese-Sign-Language\\
\hline
Number of layers &  ResNet-50 Architecture :50 resnet 8 DNN \\
& Inception-v3 Architecture: 42 inception v3 8 DNN\\
& Xception Architecture: 35 xception 8 DNN\\
& VGG-16 Architecture: 16 VGG 16 8 DNN\\

Neurons in FC layers & 1st 512, 2nd 256, 3rd 128, 4th 10 \\
Activation Function & Relu, Softmax(last layer) \\
Optimizer & Adam, Learning rate 0.0001 \\
Activation Function & Relu, Softmax(last layer) \\
Dropout & 0.5 \\
Batch size & 128 \\
Loss function & Categorical cross entropy \\
Number of epochs & 50 epochs \\
Total no.of parameters &  ResNet-50 Architecture: 2,263,178 trainable, 23,564,800 nontrainable\\
& Inception-v3 Architecture: 1,214,602 trainable, 21,802,784 nontrainable\\

& Xception Architecture: 821,386 trainable, 14,714,688 nontrainable\\
& VGG-16 Architecture: 821,386 trainable, 14,714,688 nontrainable\\
\hline
\\
\end{tabular} 
\caption{Summary of ResNet-50, Inception-v3, Xception, and VGG-16 model.} \label{modified model}
\end{table*}

To adapt the TL base model for the specific classification task, a custom classification head was added. The top/head consisted of dense layers with dropout for regularization. 

We use a sequential architecture to build a custom classification model. The following layers compose our model shown in Table \ref{modified model}:
At first,  a dense layer with neuron size of 512 units was used with the ReLU activation function. We have tried to reduce overfitting by using a dropout layer with a dropout rate of 0.4, 0.5 and 0.6 separately. Analysis the overall performance of the architecture, a dropout rate of 0.5 had outstand from other dropout layers. However to convert the output matrix of the dense layer  into a single array sized value, we used a flatten layer. Sequentially, the output of this flatten layer is passed to a dense layer of 256 units with ReLU activation function and added to another dropout layer.

Next 128 units with ReLU activation in a dense layer were used to train the output from the previous dense layer. A final dropout layer of 0.5 dropout rate is employed. At last, Softmax activation is used on a dense layer with neuron size of 10 units for classification, as there are 10 classes in our image dataset. 

Our model is compiled utilizing Adam as the optimizer and categorical cross-entropy as the loss function. We evaluated the  accuracy of our model using the evaluation metrics. InceptionV3 and VGG16 models of our proposed architecture are presented respectively in Figure \ref{DNN-inception} and Figure \ref{DNN-VGG}.

We passed the train and validation data in the model for 50 epoch and kept the batch size to 128 for each training iteration. The results are stored in the history variable for further analysis and evaluation.

 \subsubsection{Vgg16}
The VGG16 architecture is a convolutional neural network specifically developed for the purpose of picture classification. The operation of this system entails the transformation of an input image via a series of convolutional layers. The purpose of each layer is to discern increasingly complex components present in the image. Furthermore, max-pooling layers are integrated into the system in order to efficiently reduce the spatial dimensions of the image. The ultimate layers comprise of fully linked neural networks that acquire the ability to classify the image into predetermined categories. 
\subsubsection{Inceptionv3}
Google's deep convolutional neural network InceptionV3 has received a lot of praise for how it uses inception modules differently from previous versions. The network's performance in picture categorization, object identification, and other computer vision tasks is enhanced by these modules, which enable it to effectively capture diverse characteristics at different scales. For better training, we use auxiliary classifiers and batch normalization; for more precise picture classification, we use fully linked layers and global average pooling. The domains of computer vision and transfer learning greatly benefit from InceptionV3 due to its performance and adaptability.
\subsubsection{Resnet-50}
ResNet-50 is a widely used deep neural network consisting of 50 layers. It is renowned for utilizing residual blocks to enhance the training of deep networks. ResNet-50 incorporates specialized architectures to enhance computational efficiency, in addition to employing methods such as batch normalization to accelerate the learning process. The procedure concludes by utilizing global average pooling and a softmax layer for precise image classification.

% This approach has demonstrated exceptional performance on widely recognised benchmarks such as ImageNet, underscoring its importance in the field of large-scale image recognition and its applicability in transfer learning across other domains.

 \subsubsection{Xception}
Xception, short for "Extreme Inception," is a composed architecture of depthwise separable convolution layers \cite{review13}. An analysis of Inception modules in convolutional neural networks, establishing them as an intermediate stage between standard convolution and the depthwise separable convolution process, which arises the concept of Xception. Xception demonstrates a marginal enhancement in performance compared to Inception V3 when tested on the ImageNet dataset. However, it significantly outperforms Inception V3 on a bigger picture classification dataset, highlighting its efficient utilisation of parameters despite having the same number of parameters as Inception V3.

\subsubsection{Deep Neural Network} 

When it comes to modeling and extracting complicated patterns from data, a Deep Neural Network (DNN) is the artificial neural network of choice. The depth of a DNN is what sets it apart from a standard neural network; unlike standard neural networks, DNNs have numerous hidden layers between the input and output layers \cite{dnn}. Deep neural networks (DNNs) are highly effective at tasks such as image recognition, natural language processing, and more because of their ability to autonomously learn hierarchical characteristics from raw input. Each layer of a neural network contains artificial neurons that analyze data using weighted sums and activation functions. The network modifies its weights and biases via backpropagation during training to improve predictions and minimize errors.

\subsubsection{Dropout}: 
Dropout is a crucial technique in neural networks to prevent overfitting by avoiding excessive concentration on specific aspects during training. The process involves randomly deactivating certain neurons in each layer, which compels the network to learn from various viewpoints. This enhances the network's resilience and capacity to process new data, hence enhancing its overall performance. Dropout is a common technique in deep learning that enhances the efficiency of networks and improves the accuracy of outputs for various tasks.

\subsubsection{activation function}

Rectified linear activation unit (ReLU) is an essential element in the significant advancements in deep learning. It surpasses older activation functions such as sigmoid and tanh.

\begin{equation}\label{relu}
    f(x) = max(0,x)
\end{equation}

Both the original and derivative forms of the ReLU function are monotonic, as shown by the equation  \eqref{relu}. When called with a negative argument, the function prints out zero; when called with a positive argument, it prints out the positive argument. The result is an output that can take values between zero and infinity.

The softmax function, represented as $\sigma(m)$, is utilized as an activation function in deep learning for tasks involving multi-class categorization. It transforms a set of real numbers into a probability distribution that includes various classes. When assigning probabilities to various classes based on the model's output scores, the softmax function is particularly useful. Here is the softmax function's equation:

\begin{equation}\label{softmax}
\sigma(\Vec{m})_i = \frac{e^{m_i}}{\Sigma^{K}_{j=1} e^{m_j}}
\end{equation}

In equation \eqref{softmax}, $\sigma =$ softmax, $\Vec{m} =$ input vector, $e^{m_i} =$ standard exponential function for input vector, \textit{K} number of classes in the multi-class classifier,  $e^{m_j} =$ standard exponential function for output vector.

\subsubsection{loss function (categorical cross entropy)}

The goal is to avoid an extreme outlier in the greatest logit is successfully attained by label-smoothing regularization (LSR). In that case one \textit{s(k)} would approach 1 while the others converged to 0 \cite{review8}. Because unlike $s(k) = \delta_{k,y}$, all $s^{\prime}(k)$ have a positive lower bound, this would lead to a significant cross-entropy with $s^{\prime}(k)$. Cross entropy provides a different lens through which to understand LSR: The formula for the entropy of a population, $H(s^{\prime}, r)$, is as follows:

\begin{equation}\label{categorical1}
    H(s^{\prime}, r) = - \Sigma^K_{k=1} log r(k) s^{\prime} = (1 - \epsilon) H(s,r) + \epsilon H(t,r)
\end{equation}

LSR is the same as substituting two losses of cross entropy, \textit{H(s, r)} and \textit{H(t, r)}, for a single loss of cross entropy, \textit{H(s, r)} in equation \eqref{categorical1}.

The projected label distribution \textit{r} is penalized if it deviates from the prior label distribution \textit{t} by a factor of 1 in the second loss. Considering that $H(t, r) = DKL(tkr) + H(t)$, where \textit{H(t)} is a constant, it is clear that the KL divergence may capture this deviation. \textit{H(t, r)} is a measure of how different the anticipated distribution \textit{r} is from the uniform distribution \textit{t}, which might alternatively be assessed (though not equivalently) by negative entropy \textit{H(r)}.

\subsubsection{Optimizer ADAM}
The Adam (Adaptive Moment Estimation)  optimizer is employed for training deep neural networks. This approach combines the advantageous aspects of two prominent optimisation techniques, specifically RMSprop and Momentum. Adam is well-known for its ingenious method of learning rate adjustment, which ensures a more rapid and consistent convergence to the optimal solution. To achieve this, it integrates concepts from RMSprop and Momentum. In essence, it gains forward motion and modifies the magnitude of its steps in response to gradients in the past. Additionally, it checks for biases to make sure it isn't overly biased in one direction, particularly when training first begins. Because of its versatility, Adam is a great tool for neural network training in general.

\subsubsection{SHAP} 
The SHapley Additive exPlanations (SHAP) approach has implications in the field of explainable AI. It utilizes concepts from game theory to elucidate the process by which machine learning models generate predictions. SHAP aids in simplifying intricate model decisions by illustrating the significance of each feature in predicting an outcome. This enhances the comprehensibility of models, aids in error detection, uncovers biases, and instills confidence in AI systems. It is utilized across several sectors such as healthcare, banking, and language processing to enhance the transparency and reliability of AI.
Its versatility and effectiveness make it an indispensable tool for professionals and individuals involved in understanding and enhancing the behaviour of machine learning models.

\begin{figure}[h!]
\centerline{\includegraphics[width=3.5in, height=1.48in]{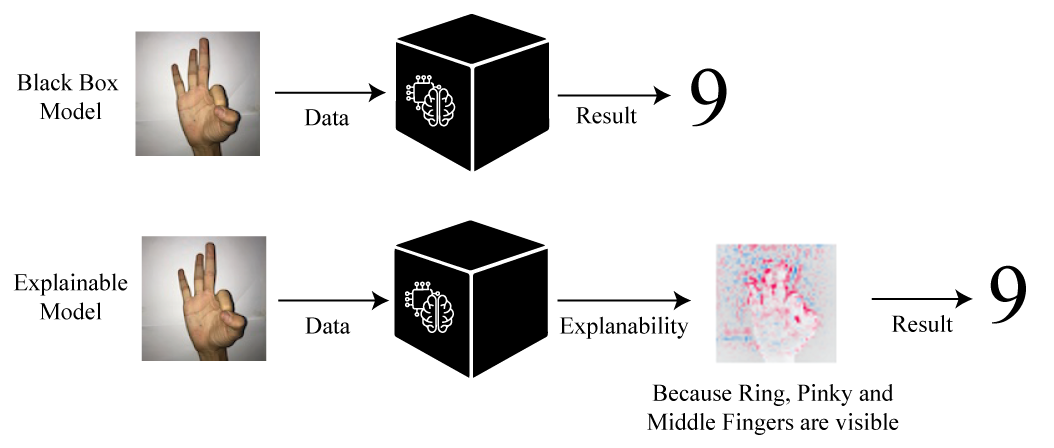}}

\caption{SHAP}
\label{SHAP}
\end{figure}

Figure \ref{SHAP} gives insight into the workings of our models' Explainability. The operation of our first machine learning model is opaque, which resembles a black-box model. However, after the application of Explainability, clarity emerges to discern the rationale behind our model's results. This transition enables a deeper understanding of the decision-making process, renouncing light on the previously concealed mechanisms within our architecture.

\section{Evaluation Metrics}\label{eva matrix}
In addition to these two metrics, we also used the precision and recall as well as the F1-score to test each of our modified architectures separately for the 10 classes of Bhutanese numerical sign. 

Precision is synonymous with the positive predictive value. Prediction accuracy is measured as the ratio of predicted accurate class values to the total number of predictions. The precidion is computed with the help of the equation \eqref{precision equation}. 

\begin{equation}\label{precision equation}
    Precision = \frac{True\_Positive}{True\_Positive + False\_Positive}
\end{equation}

One definition of recall is the fraction of right predictions as a fraction of the total number of correct class values. The equation \eqref{recall equation} is used to calculate the recall. 

\begin{equation}\label{recall equation}
    Recall = \frac{True\_Positive}{True\_Positive + False\_Negative} 
\end{equation}

The F-score, sometimes known as the F-measure, is an often used statistical tool. The F1-score reflects the balance between precision and recall. The F1-score improves significantly only in the presence of high precision and recall values. F1-score values range from 0 to 1, with higher values indicating more precise classification. The equation \eqref{f1 equation} is used to determine the F1-score.

\begin{equation}\label{f1 equation}
    F1-score = \frac{2 \ast Precision \ast Recall}{Precision + Recall} 
\end{equation}

\section{Result and Discussion}\label{result}
The hand sign language recognition system's performance is assessed using a split dataset's test set. The test set comprises 20,000 images, with a distribution of 70\% for the training set, 15\% for the validation set, and 15\% for the test set. This distribution equates to around 14,000 samples for training, 3,000 samples for validation, and 3,000 samples for testing. The evaluation is based on these 6,000 samples. The accuracies of the two models, Inceptionv3 and ResNet50, ranged from 95.47\% to 98.90\%. We may infer that the model effectively addresses overfitting during training and attains a test accuracy that closely mirrors the training accuracy, as there is a clear and consistent positive correlation between training accuracy and testing accuracy throughout the 50 epochs.

%The left side of Figure \ref{Inception-v3 Training and Validation Loss and Accuracy} depicts the accuracy of the training versus the accuracy of the validation across 50 epochs for the Inceptionv3 model, while the right side indicates the loss of the validation across 50 epochs for the Inceptionv3 model. Then, Figure \ref{ResNet-50 Training and Validation Loss and Accuracy} shows the accuracy of the training versus the accuracy of the validation across 50 epochs for the ResNet50 model, as well as the validation loss across 50 epochs for the ResNet50 model.

% \begin{figure}[t]
% \centerline{\includegraphics[width=3.5in, height=1.76in]{images/inceptionv3-trainingand_validation_loss_and_accuracy.png}}

% \caption{Inception-v3 Training and Validation Loss and Accuracy}
% \label{Inception-v3 Training and Validation Loss and Accuracy}
% \end{figure}

% \begin{figure}[t]
% \centerline{\includegraphics[width=3.5in, height=1.76in]{images/resnet 50 training and validation loss and accuracy.png}}

% \caption{ResNet-50 Training and Validation Loss and Accuracy}
% \label{ResNet-50 Training and Validation Loss and Accuracy}
% \end{figure}

\begin{figure}[t]
\centerline{\includegraphics[width=3in, height=2.5in]{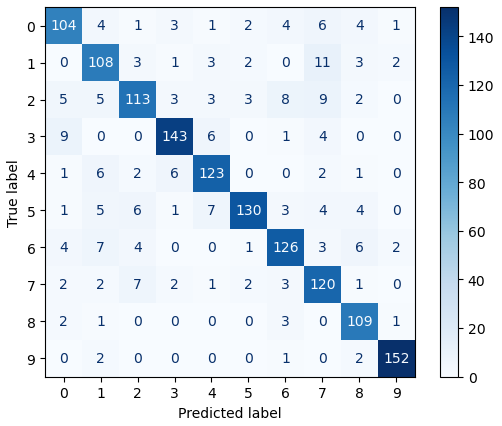}}

\caption{Inceptionv3 Confusion Matrix}
\label{Inceptionv3 Confusion Matrix}
\end{figure}

\begin{figure}[t]
\centerline{\includegraphics[width=3in, height=2.5in]{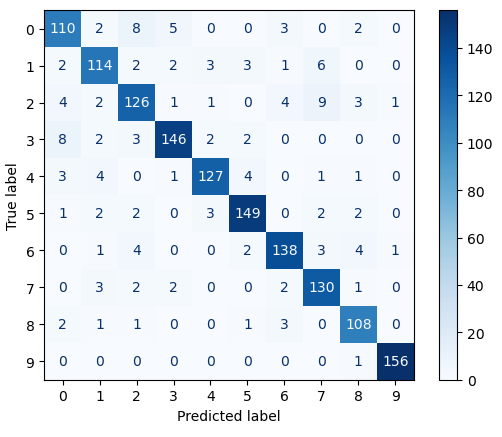}}

\caption{ResNet-50 Confusion Matrix}
\label{ResNet-50 Confusion Matrix}
\end{figure}

The confusion matrix provides additional metrics to assess the performance of our model for each class separately. Confusion matrices, as shown in Figure \ref{Inceptionv3 Confusion Matrix}-\ref{ResNet-50 Confusion Matrix}, are used to evaluate the outcomes of projections. The number of accurate and inaccurate estimations are counted and grouped by category. The confusion matrix is a 10$\ast$10 matrix representing the classification results for 10 classes. The white cells in the matrix represent a low cell value, whereas the blue cells represent a higher cell value.

An overview of our findings with various models is illustrated in Table \ref{table precision}. The table presents a summary of the results based on the metrics previously discussed. The combined metric, F1-score, indicates that our suggested models perform well by accurately classifying all classes in the test set. 
With Vgg16, we were able to get a 98.53\% training accuracy, 92.04\% test accuracy, 92.33\% precision, 92.04\% F1 score, and 92.04\% recall. The training accuracy, test accuracy, recall, F1 score, and precision of ResNet50 are 99.99 percent, 90.24\%, 90.24\%, and 90.24\%, respectively. The metrics we measured with Inceptionv3 model are as follows: training accuracy of 95.47 percent, test accuracy of 84.98\%, precision of 84.85\%, F1 score of 84.98\%, and recall of 84.98\%. The training accuracy rates, test accuracy rates, precision rates, F1 score, and recall rates of Xception were 98.80\%, 87.75\%, 87.59\%, and 87.59\%, respectively.

\begin{table}[t]

\caption{Model accuracy} \label{table precision}
\begin{tabular}
{ |p{1.45cm}|p{1cm}|p{1cm}|p{1.15cm}|p{.9cm}| p{.9cm}|p{1cm}|}

 \hline

 \hline
Algorithm&  Training  Accuracy & Test  Accuracy & Precision&   F1 Score &   Recall\\
 \hline
  Vgg16   & 98.53\%    &92.04\% &  92.33\%&  92.04\%&   92.04\%\\
 ResNet50 & 98.90\%    & 90.24\% &  90.04\% &  90.24\% & 90.24\%\\
 Inceptionv3 & 95.47\%    &84.98\%&  84.85\%&84.98\% &84.98\%\\
 
 Xception   & 98.80\%    &87.75\%&  87.59\%&  87.75\% &87.75\%\\
 \hline
\end{tabular}
\end{table}

\subsection{Explainability of the Model}

To enhance the interpretability of our model, we incorporated explainability strategies in our research work. This study employed the SHAP deep explanation, a technique based on cooperative game theory, to assess the influence of individual features on the machine learning model's output. This strategy offered valuable insights into the importance of features in intricate models.

\begin{figure*}[t]
\centerline{\includegraphics[width=6in, height=3.4in]{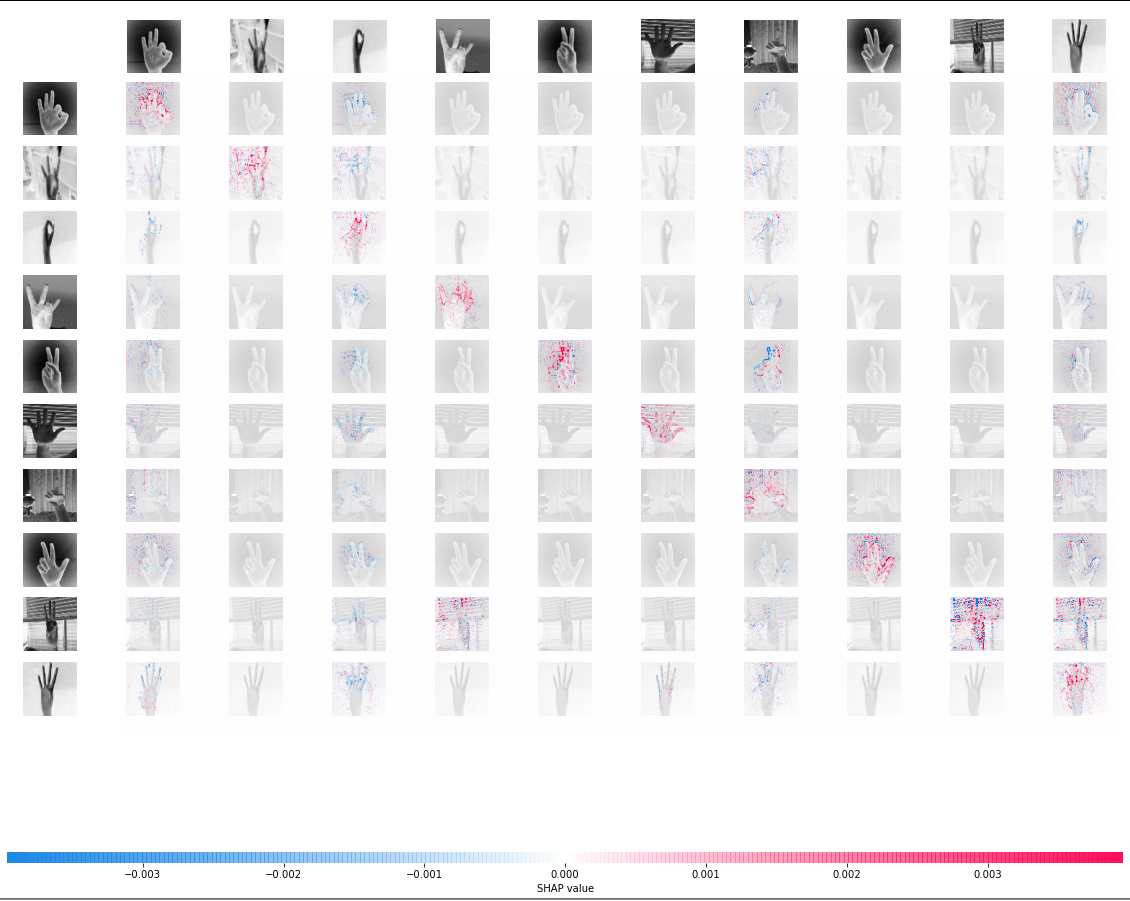}}

\caption{Explainability of the Model}
\label{Explainability of the Model}
\end{figure*}

The effectiveness of the SHAP deep explainer can be seen by the visual representation. In Figure \ref{Explainability of the Model}, pixels were color-coded according to their value in classifying certain categories. Pixels with higher importance were emphasized in pink, while those with lower importance were highlighted in blue. The vertical axis depicts the class designations, and the horizontal axis reflects the level of resemblance.

The visualization presented in Figure \ref{Explainability of the Model} demonstrates that the model assigns significant weight to edge pixels surrounding the hand in all categories, particularly when it comes to features related to the fingers.

\begin{table*}[t]
\centering
\begin{tabular}{c  c  c  c}
\hline
Reference & Dataset & Feature extraction & Accuracy \\
\hline
\cite{review1}& Thomas Moeslund’s gesture recognition dataset & Transfer Learning Based CNN & around 98\% \\
\cite{review2} &ArSL2018  & ResNet50 and MobileNetV2 & 97\% \\ 
\cite{review5} & & Capsnet & 94.2\% \\
\cite{review7} & MUGD, ISL, ArSL, NUS-I, and NUS-II & Inception v3 & above 90\% \\
\cite{review12} & Intel Image Classification Challenge dataset & Xception & around 90\% \\
Our model & Bhutanese-Sign-Language (BSL) & Vgg16  & 98.53\% \\
& &ResNet50 & 98.90\% \\
& &Inceptionv3 & 95.47\% \\
& &Xception & 98.80\% \\
\hline
\\
\end{tabular}

\caption{Comparison of related work}\label{compare}
\end{table*}

After conducting a more thorough analysis of the figure, it was observed that during the prediction of the '1' class, the model attributed the greatest significance to the index finger, while assigning relatively lower importance to the area associated with the little finger.

\section{Comparative Analysis}\label{Comparative Analysis}

Researchers have dedicated significant work to developing methods for facilitating communication between individuals who are proficient in speech but unfamiliar with sign language and those who are deaf or hearing-impaired and predominantly utilize sign language. Each culture possesses a unique sign language that differs from other civilizations, leading to countless research being undertaken worldwide across different countries and cultures. Here, we will take a close look at studies that have examined Arabic sign language. Several recent research were reviewed because of their close relevance to our own. Table \ref{compare} displays the results of some of these investigations.

\section{Conclusion}\label{conclude}

%This study presents the development of a sign language recognition system using a Deep Neural Network (DNN). In order to get strong and reliable performance, we utilised the Resnet-50, Inception-v3, Xception, and VGG-16 models for sign language recognition. We carried out investigations to train and test on Bhutanese Sign Language. Additionally, we examined the comprehensibility of the provided data with the SHAP. Most of the current study focused on the result rather than explainability. Future research could potentially involve the integration of pretrained models to improve performance and provide better explanations.

This paper explores the complexities involved in creating a Deep Neural Network (DNN)-based sophisticated sign language recognition system. The models Resnet-50, Inception-v3, Xception, and VGG-16 were specifically selected to guarantee the robustness and dependability of the system in correctly deciphering sign language gestures. Notably, the inquiry focused on the distinctive linguistic aspects of Bhutanese Sign Language during both the training and assessment phases. Additionally, the study went beyond a simple performance assessment and examined comprehension by using the SHAP (SHapley Additive exPlanations) technique to analyse and comprehend the underlying patterns in the data that were supplied. Although obtaining favourable findings was the main focus of the current study, explainability in model predictions is becoming increasingly important. The paper makes recommendations for potential directions for future research that could use pretrained model integration. In addition to enhancing the system's overall functionality, this strategic improvement seeks to give the DNN's recognition judgements more understandable and transparent justifications. More complex and inclusive communication technologies are being made possible by the dynamic junction of interpretability and accuracy in sign language recognition systems. Expanding upon our prior investigation, our objective is to further examine the interpretability of the model's forecasts through an exploration of the distinct characteristics or recurring patterns detected by the SHAP deep explainer across various sign language gestures. Furthermore, we shall examine the possible ramifications that may arise from the integration of user feedback mechanisms into the system.

%This study delves into the intricacies of developing a sophisticated sign language recognition system through the implementation of a Deep Neural Network (DNN). The Resnet-50, Inception-v3, Xception, and VGG-16 models were carefully chosen to ensure the system's strength and reliability in accurately interpreting sign language gestures. Notably, the investigation focused on the unique linguistic characteristics of Bhutanese Sign Language during both the training and testing phases. Furthermore, the study went beyond mere performance evaluation and delved into the comprehensibility aspect by employing the SHAP (SHapley Additive exPlanations) technique to interpret and understand the underlying patterns in the provided data. While the primary emphasis of the current study was on achieving favorable results, there is a recognition of the growing importance of explainability in model predictions.

%Looking ahead, the study suggests future research avenues that could involve the integration of pretrained models. This strategic enhancement aims not only to improve the system's overall performance but also to provide more transparent and insightful explanations for the recognition decisions made by the DNN. The evolving intersection of accuracy and interpretability in sign language recognition systems stands as a key focal point, paving the way for more advanced and inclusive communication technologies.

\end{document}